\documentclass{article}
\usepackage{spconf,amsmath,graphicx}
\usepackage{hyperref}
\usepackage{subcaption}
\usepackage{graphicx}

\usepackage{enumitem}
\setlist{nosep, leftmargin=14pt}

\usepackage{mwe} 

\title{Localizing and Assessing Node Significance in Default Mode Network using Sub-Community Detection in Mild Cognitive Impairment}

\name{Ameiy Acharya$^{\star}$ \qquad Chakka Sai Pradeep $^{\dagger}$ \qquad Neelam Sinha$^{\star}$}

\address{$^{\star}$Centre for Brain Research, Indian Institute of Science, Bengaluru \\
         $^{\dagger}$International Institute of Information Technology, Bangalore}

\begin{document}
%
\maketitle
\begin{abstract}
Our study aims to utilize fMRI to identify the affected brain regions within the Default Mode Network (DMN) in subjects with Mild Cognitive Impairment (MCI), using a novel Node Significance Score (NSS). We construct subject-specific DMN graphs by employing partial correlation of Regions of Interest (ROIs) that make-up the DMN. For the DMN graph, ROIs are the nodes and edges are determined based on partial correlation. Four popular community detection algorithms (Clique Percolation Method (CPM), Louvain algorithm, Greedy Modularity and Leading Eigenvectors) are applied to determine the largest sub-community. NSS ratings are derived for each node, considering (I) frequency in the largest sub-community within a class across all subjects and (II) occurrence in the largest sub-community according to all four methods. After computing the NSS of each ROI in both healthy and MCI subjects, we quantify the score disparity to identify nodes most impacted by MCI. The results reveal a disparity exceeding 20\% for 10 DMN nodes, maximally for PCC and Fusiform, showing 45.69\% and 43.08\% disparity. This aligns with existing medical literature, additionally providing a quantitative measure that enables the ordering of the affected ROIs. These findings offer valuable insights and could lead to treatment strategies aggressively targeting the affected nodes.


\end{abstract}
\section{Introduction}
\label{sec:intro}

Among the elderly population, the risk of neurodegeneration leading to Alzheimer's Disease (AD) stands at 10\% after the age of 65 rising to 40 \% after the age of 85 \cite{kumar2022alzheimer}. The beginning of this cognitive decline manifests as Mild Cognitive Impairment (MCI), a stage in which the neurodegeneration can be potentially reversed if detected early and treated aggressively. Thus it is important to identify and understand the various aspects of MCI. Resting state Functional MR imaging (rs-fMRI) is the tool of choice in studying degeneration since it allows Default Mode Network (DMN) to be studied carefully. Medical literature points to discrepancies in DMN in neurodegeneration. There are myriad aspects of the DMN that need to be studied in order to arrive at a holistic inference.

\begin{figure*}[ht]
\centering
\includegraphics[width=\linewidth]{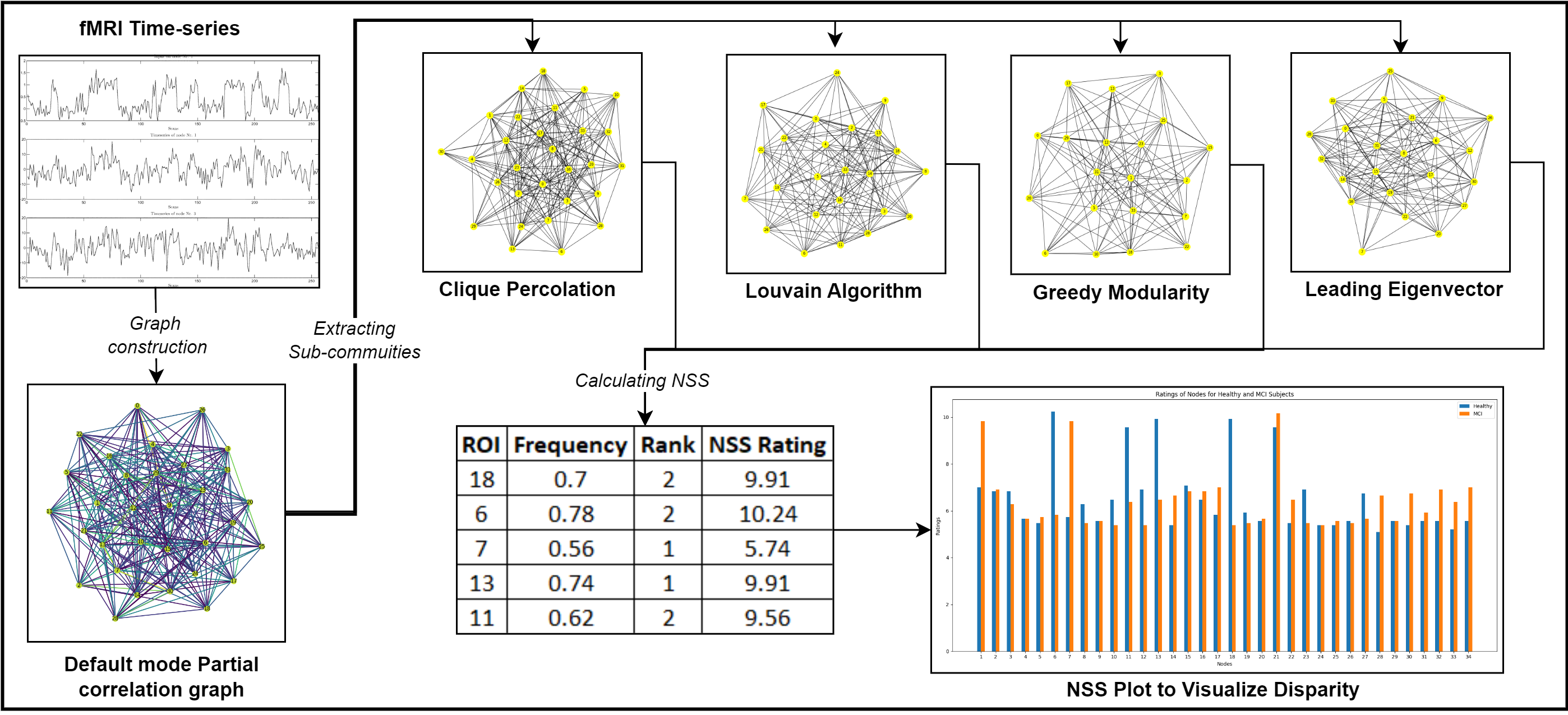}
\caption{Overview of proposed graph-based method}
\label{fig:MethodBlock}
\end{figure*}

\section{Literature survey}
\label{sec:Literature survey}

Functional brain networks are widely explored using graph-based techniques. Existing literature covers various aspects of the graph framework in the context of functional brain networks. Some studies focus on static graphs representing cognitive states and subsequent analysis, while others emphasize the importance of understanding graph evolution over time. Although many studies use small datasets, some address scalability concerns.

In \cite{zhang}, authors combine proximal relations and semantic connections through deep graph hashing learning for classification and identifying abnormalities. In \cite{zhu2023classification}, deep graph convolutional neural networks are applied to large multi-site datasets for classifying brain networks related to major depressive disorders. \cite{Junzhong} introduces a technique, that employs a hypergraph attention network to represent high-order structural information in brain networks.

Community detection is explored in \cite{Garcia_CommDetect} to investigate small-world properties and modularity in neuroimaging data. \cite{bi2020functional} focuses on brain network classification for Alzheimer's disease detection, utilizing recurrent learning for deep adjacent positional features and deep regional-connectivity features. An Extreme Learning Machine (ELM)-boosted structure is proposed to enhance performance.

These studies collectively contribute to our understanding of functional brain networks, covering a range of methods and applications.
In the proposed work, the idea of the largest sub-community of the graph as holding information about the entire graph is hypothesized. The largest sub-community of a graph happens to be that region of the graph where maximum number of nodes interact and influence each other. It is conceivable that in the event of neurodegeneration the ability of the nodes to interact and influence one another is potentially hampered. Thus identification of the largest sub-community of the DMN could lead to understanding some aspects of regions that influence one another in a healthy brain.
\section{Methodology}
\label{sec:Methodology}
The study aims at identifying regions that are significant in understanding MCI. Towards this, contributions of the study are:
\begin{itemize}
    \item Localizing structures responsible for neurodegeneration in MCI using existing approaches in Graph-based sub-community detection
    \item Proposed ``Node Significance score" to measure participation of nodes in a cognitive network
    \item Quantified the disparity in the proposed NSS for all nodes across classes
\end{itemize}
The proposed method as seen in Fig \ref{fig:MethodBlock} involves building a graph of the regions that participate in the DMN, using rs-fMRI time series of the respective voxel locations. The constructed graph is analyzed using four popular sub-community detection methods that look for the largest sub-community. A novel measure to compute node participation that factors the following two aspects: (I) Presence of a node in largest sub-community, across all four sub-community detection methods (II) Consistency across subjects

\subsection{Partial Correlation Measure}
The statistical measure, ``Partial correlation between nodes $i$, $j$ denoted as $\rho_{ij}$", which assesses correlations while regressing out effects of other nodes, is used to construct a graph, using time series of voxel intensities for each DMN ROI. When $\rho_{ij}$ is zero, nodes $i$ and $j$ are conditionally independent given other nodes, making it a robust measure of direct connectivity. Partial correlation, given precision matrix $\omega$, is computed as $\rho_{ij} = \frac{-\omega_{ij}}{\sqrt{\omega_{ii}\omega_{jj}}}$. We use Moore-Penrose pseudo-inverse to estimate partial correlation.

\textbf{Graph Construction:} Subject-specific undirected graphs were built with partial correlations as edge weights. Each ROI represents a node, and edges are established based on partial correlation values. These weighted graphs reflect the strength and nature of connectivity between DMN ROIs.

\subsection{Data Used}
The rs-fMRI time series volumes of 50 MCI and 50 healthy subjects are taken from publicly available  Alzheimer's Disease Neuroimaging Initiative (ADNI) database \cite{adni1}. Data was processed using FSL version 6.0.4 \cite{jenkinson2012fsl}. The standard fMRI preprocessing steps of motion correction, slice timing adjustment, normalization to MNI space, and nuisance variable regression, were carried out.

\subsection{Sub-Community Detection}

The Clique Percolation Method (CPM) introduced in \cite{CPM} is employed for identifying overlapping sub-communities in a graph. CPM is based on the concept of k-cliques, fully connected subgraphs of size k. This iterative technique involves detecting k-cliques and merging them to form new sub-communities. 

Modularity is a function that measures the extent to which nodes in a network are grouped into sub-communities. This is used in Louvain algorithm and Greedy Modularity. This is defined as a value in the range $[-1/2,1]$ that measures the density of links within sub-communities compared to links across sub-communities. For a weighted graph, modularity is defined as:

\begin{equation}
Q = \frac{1}{2m}\sum\limits_{ij}\left[A_{ij} - \frac{k_i k_j}{2m}\right]\delta (c_i,c_j),
\end{equation}

Where $A_{ij}$ represents the edge weight between nodes $i$ and $j$; $k_i$ and $k_j$ are the sum of the weights of the edges attached to nodes $i$ and $j$, respectively; $m$ is the sum of all edge weights in the graph; $c_i$ and $c_j$ are the communities of the nodes; and $\delta$ is the Kronecker delta function.

The Louvain algorithm, introduced in \cite{Louvain}, optimizes modularity. We applied the Louvain algorithm to partition the network into non-overlapping sub-communities and reveal the network's hierarchical structure. On the other hand, Greedy Modularity Optimization method introduced in \cite{chen2014community} also maximizes modularity, identifying cohesive, non-overlapping sub-communities in the graph. Furthermore, the Leading Eigenvector Method \cite{newman2006finding}, known as spectral clustering, leverages spectral properties of the graph's adjacency matrix to partition the graph using the eigenvector corresponding to the second smallest eigenvalue. We apply all four techniques on DMN graphs and obtain the respective largest sub-community for each subject across all the methods. 

\subsection{Node Significance Score (NSS)} 
\begin{figure*}[ht]
\centering
\includegraphics[width=\linewidth]{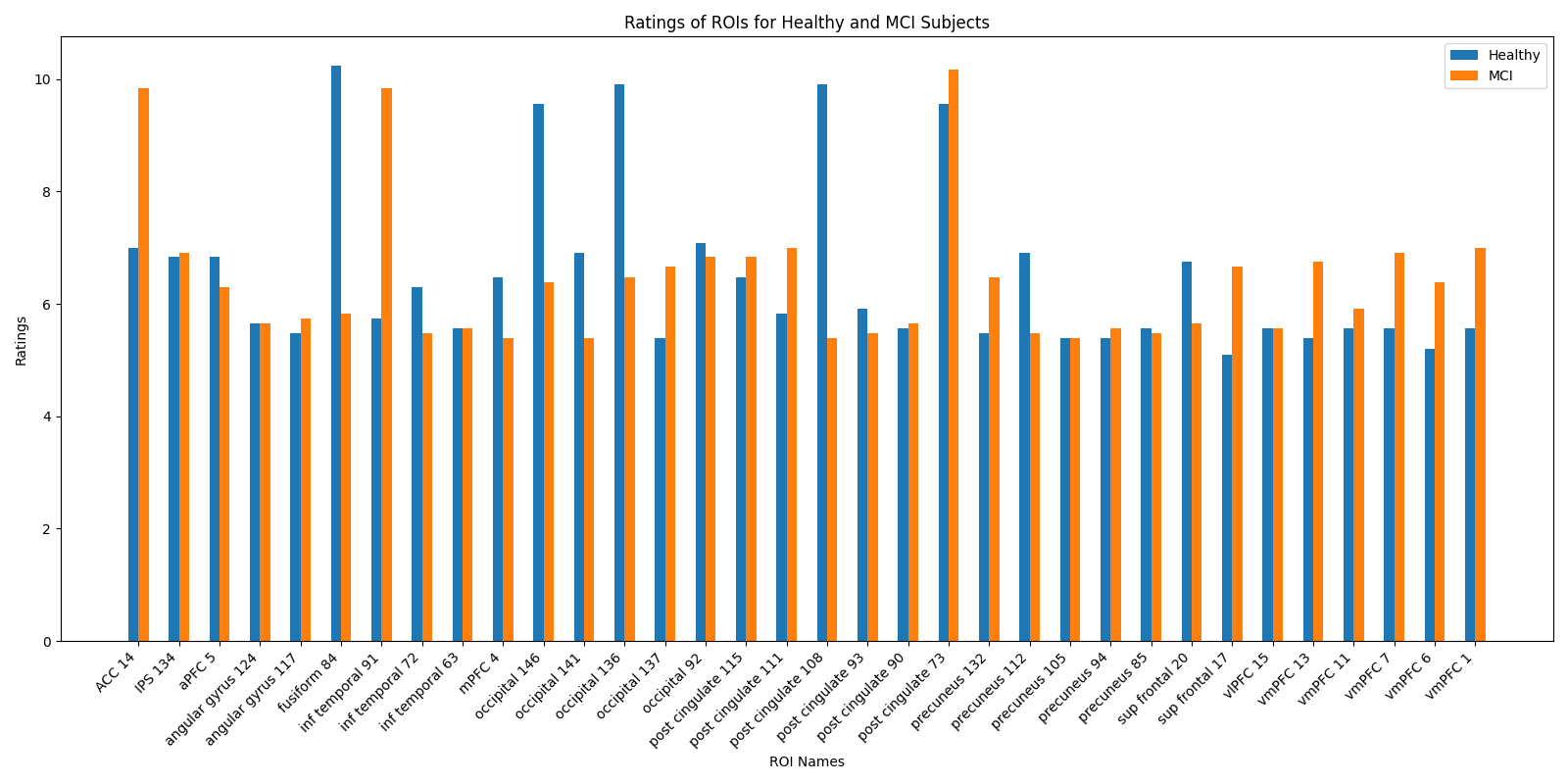}
\caption{Comparison of Proposed Node Significance score (NSS) for ROIs in DMN}
\label{fig:stream}
\end{figure*}

This computation comprises of two parts- (I) ``Highest occurrence of a given ROI" which is determined as the fraction of subjects in which a given ROI is detected in the largest sub-community, irrespective of the sub-community detection method. (II) We propose an intermediate ``node rank" as the number of methods that identify the node in the largest sub-community, ranging from 4 (identified by all) to 1 (identified by only one). Utilizing (I) and (II),  we propose Node Significance Score (NSS) as:
\begin{equation}
    \textit{N} = \textit{r}^2 + \sqrt{\textit{h}} , 
\end{equation}
where $N$, $r$ and $h$ are the NSS score, node rank and Highest occurrence of a given ROI, respectively.
ROI-specific NSS for both healthy and MCI subjects is plotted in Fig \ref{fig:stream}. Percentage disparity in NSS between MCI and healthy subjects for each ROI is tabulated in Table \ref{Table1}. The code for the proposed methodology is provided on GitHub: \url{https://tinyurl.com/4bkumw4w}.


\section{Results and Discussion}
\label{sec:Results and Discussion}

\begin{table}[ht]
\centering
\begin{tabular}{|c|c|}
\hline
ROI Name & Percentage Difference \\
\hline
post cingulate 108 & 45.69 \\
fusiform 84 & 43.08\\
inf temporal 91 & 41.56 \\
occipital 136 & 34.67 \\
occipital 146 & 33.26\\
\hline
\end{tabular}
\caption{Disparity in proposed Node Significance score (NSS) between Healthy and MCI subjects, for Top-5 nodes}
\label{Table1}
\end{table}
The objective of the investigation is to identify those nodes that show differential behavior across the subject groups. This enables ordering in affected ROIs and also could lead to targeted treatment strategies. Peak disparity of 45.69\% is seen in PCC-108, followed by 43.08\% for Fusiform-84. In all, 10 nodes show disparity of over 20\%. Hence this set of nodes are identified to be vital to understand MCI.


 \textbf{Biological Significance:} The Posterior Cingulate Cortex (PCC) has been identified as a crucial region within the DMN with the highest number of connections \cite{raichle2001default}. Extensive medical literature supports the pivotal role of PCC in various cognitive functions, including memory retrieval, attention, salience, and emotion \cite{leech2014role, mortamais2017detecting}.  Furthermore, the identification of Fusiform-84 as another significant region aligns with previous research \cite{Fusiform} showcasing its decline in MCI. 
 
 Notably, our method goes beyond merely pointing out the importance of specific ROIs; it quantifies the disparity caused by neurodegeneration, providing another dimension to understand its significance. This quantification allows for the precise targeting of regions, a valuable insight for advanced methods like connectomes. Generalizability of our method makes it applicable to other functional connectivity networks such as sensorimotor and frontoparietal.

\textbf{Comparing Algorithms:} While CPM identifies communities through percolation of overlapping k-cliques, Louvain and Greedy Modularity techniques primarily identify non-overlapping sub-communities by optimizing modularity, making the analysis exhaustive. Leading Eigenvectors relies on spectral analysis of Adjacency matrix to identify non-overlapping communities. Hence the four different methods lead to comprehensive understanding of DMN.

\section{Conclusion}
\label{sec:Conclusion}
In this work, we investigate regions responsible for neurodegeneration in MCI, using graph-based approaches. The hypothesis is that the regions most affected in MCI would show different characteristics when compared against healthy controls. The behavior studied here is, level of participation in DMN, in the two subject groups (balanced dataset of 100 subjects sourced from ADNI). Utilizing four popular sub-community detection methods, as well as, factoring the consistency across subjects, we propose a measure called ``Node Significance Score" (NSS) to identify the activity level of each node. Results show, peak disparity in NSS to be 45.69\% for PCC-108, followed by 45\% for Fusiform-84. Additionally, 10 nodes in DMN exhibit disparity of more than 20\%. This illustrates promise in the proposed approach towards understanding neurodegeneration in MCI. This method could also be extended to other aspects of cognitive understanding.

\bibliographystyle{IEEEbib}
\bibliography{strings,refs}

\begin{thebibliography}{10}

\bibitem{kumar2022alzheimer}
A~Kumar, J~Sidhu, A~Goyal, and JW~Tsao,
\newblock ``Alzheimer disease. 2021 aug 11,''
\newblock {\em StatPearls. Treasure Island (FL): StatPearls Publishing}, 2022.

\bibitem{zhang}
Junzhong Ji and Yaqin Zhang,
\newblock ``Functional brain network classification based on deep graph hashing learning,''
\newblock {\em IEEE Transactions on Medical Imaging}, vol. PP, pp. 1--1, 05 2022.

\bibitem{zhu2023classification}
Manyun Zhu, Yu~Quan, and Xuan He,
\newblock ``The classification of brain network for major depressive disorder patients based on deep graph convolutional neural network,''
\newblock {\em Frontiers in Human Neuroscience}, vol. 17, pp. 1094592, 2023.

\bibitem{Junzhong}
Junzhong Ji, Yating Ren, and Minglong Lei,
\newblock ``Fc–hat: Hypergraph attention network for functional brain network classification,''
\newblock {\em Information Sciences}, vol. 608, pp. 1301--1316, 2022.

\bibitem{Garcia_CommDetect}
Javier~O. Garcia, Arian Ashourvan, Sarah Muldoon, Jean~M. Vettel, and Danielle~S. Bassett,
\newblock ``Applications of community detection techniques to brain graphs: Algorithmic considerations and implications for neural function,''
\newblock {\em Proceedings of the IEEE}, vol. 106, no. 5, pp. 846--867, 2018.

\bibitem{bi2020functional}
Xin Bi, Xiangguo Zhao, Hong Huang, Deyang Chen, and Yuliang Ma,
\newblock ``Functional brain network classification for alzheimer’s disease detection with deep features and extreme learning machine,''
\newblock {\em Cognitive Computation}, vol. 12, pp. 513--527, 2020.

\bibitem{adni1}
Clifford~R Jack~Jr, Matt~A Bernstein, et~al.,
\newblock ``The \uppercase{A}lzheimer's disease neuroimaging initiative (\uppercase{ADNI}): \uppercase{MRI} methods,''
\newblock {\em Journal of Magnetic Resonance Imaging: An Official Journal of the International Society for Magnetic Resonance in Medicine}, vol. 27, no. 4, pp. 685--691, 2008.

\bibitem{jenkinson2012fsl}
Mark Jenkinson, Christian~F Beckmann, Timothy~EJ Behrens, Mark~W Woolrich, and Stephen~M Smith,
\newblock ``Fsl,''
\newblock {\em Neuroimage}, vol. 62, no. 2, pp. 782--790, 2012.

\bibitem{CPM}
Gergely Palla, Imre Derényi, Illés Farkas, and Tamás Vicsek,
\newblock ``Uncovering the overlapping community structure of complex networks in nature and society,''
\newblock {\em Nature}, vol. 435, pp. 814--818, 07 2005.

\bibitem{Louvain}
Vincent Blondel, Jean-Loup Guillaume, Renaud Lambiotte, and Etienne Lefebvre,
\newblock ``Fast unfolding of communities in large networks,''
\newblock {\em Journal of Statistical Mechanics Theory and Experiment}, vol. 2008, 04 2008.

\bibitem{chen2014community}
Mingming Chen, Konstantin Kuzmin, and Boleslaw~K Szymanski,
\newblock ``Community detection via maximization of modularity and its variants,''
\newblock {\em IEEE Transactions on Computational Social Systems}, vol. 1, no. 1, pp. 46--65, 2014.

\bibitem{newman2006finding}
Mark~EJ Newman,
\newblock ``Finding community structure in networks using the eigenvectors of matrices,''
\newblock {\em Physical review E}, vol. 74, no. 3, pp. 036104, 2006.

\bibitem{raichle2001default}
Marcus~E Raichle, Ann~Mary MacLeod, Abraham~Z Snyder, William~J Powers, Debra~A Gusnard, and Gordon~L Shulman,
\newblock ``A default mode of brain function,''
\newblock {\em Proceedings of the national academy of sciences}, vol. 98, no. 2, pp. 676--682, 2001.

\bibitem{leech2014role}
Robert Leech and David~J Sharp,
\newblock ``The role of the posterior cingulate cortex in cognition and disease,''
\newblock {\em Brain}, vol. 137, no. 1, pp. 12--32, 2014.

\bibitem{mortamais2017detecting}
Marion Mortamais, Jessica~A Ash, John Harrison, Jeffrey Kaye, Joel Kramer, Christopher Randolph, Carine Pose, Bruce Albala, Michael Ropacki, Craig~W Ritchie, et~al.,
\newblock ``Detecting cognitive changes in preclinical alzheimer's disease: A review of its feasibility,''
\newblock {\em Alzheimer's \& dementia}, vol. 13, no. 4, pp. 468--492, 2017.

\bibitem{Fusiform}
Arun Bokde, P~Lopez-Bayo, T~Meindl, S~Pechler, C~Born, Frank Faltraco, Stefan Teipel, H.-J Möller, and Harald Hampel,
\newblock ``Functional connectivity of the fusiform gyrus during a face-matching task in subjects with mild cognitive impairment,''
\newblock {\em Brain : a journal of neurology}, vol. 129, pp. 1113--24, 06 2006.

\end{thebibliography}

\end{document}